\documentclass[letterpaper, 10 pt, conference]{ieeeconf}  

\IEEEoverridecommandlockouts                              

\usepackage{graphics} 
\usepackage{epsfig} 
\usepackage{times} 
\usepackage{amsmath} 
\usepackage{amssymb}  
\usepackage{hyperref}
\usepackage{newtxtext,newtxmath}
\usepackage{makecell}
\usepackage{multirow}

\title{\LARGE \bf
3D Affordance Keypoint Detection for Robotic Manipulation
}

\author{Zhiyang Liu$^{1}$, Ruiteng Zhao$^{1}$, Lei Zhou$^{1}$, Chengran Yuan$^{1}$, Yuwei Wu$^{2}$, Sheng Guo$^{1}$,\\ Zhengshen Zhang$^{1}$, Chenchen Liu$^{1}$ and Marcelo H Ang Jr$^{1}$
\thanks{$^{1}$Advanced Robotics Centre, National University of Singapore, 117576, Singapore, \{zhiyang, ruiteng, leizhou, chengran.yuan, e0576004, zhengsheng\_zhang, chenchen.liu\}@u.nus.edu, mpeangh@nus.edu.sg}%
\thanks{$^{2}$Department of Mechanical Engineering, National University of Singapore, 117575, Singapore}%
}

\begin{document}

\maketitle
\thispagestyle{empty}
\pagestyle{empty}

\begin{abstract}

This paper presents a novel approach for affordance-informed robotic manipulation by introducing 3D keypoints to enhance the understanding of object parts' functionality. The proposed approach provides direct information about \textit{what} the potential use of objects is, as well as guidance on \textit{where} and \textit{how} a manipulator should engage, whereas conventional methods treat affordance detection as a semantic segmentation task, focusing solely on answering the \textit{what} question. To address this gap, we propose a Fusion-based Affordance Keypoint Network (FAKP-Net) by introducing 3D keypoint quadruplet that harnesses the synergistic potential of RGB and Depth image to provide information on execution position, direction, and extent. Benchmark testing demonstrates that FAKP-Net outperforms existing models by significant margins in affordance segmentation task and keypoint detection task. Real-world experiments also showcase the reliability of our method in accomplishing manipulation tasks with previously unseen objects.

\end{abstract}

\section{INTRODUCTION}

Autonomous robotic manipulation requires robots to understand the various potential functions of objects and this understanding is referred to as "affordance" \cite{gibson1977theory}. Unlike other properties such as object pose that solely describes the object itself, affordances consider the functional interactions between an object's parts and humans or robots \cite{hart_affordance_2015}. According to recent studies, the affordance detection for object parts has been approached as a semantic segmentation problem \cite{hadjivelichkov_one-shot_2023,chu_toward_2019,chu_learning_2019,do_affordancenet_2018,nguyen_object-based_2017,nguyen_detecting_2016,sawatzky_weakly_2017,myers_affordance_2015} where affordances are predicted by grouping pixels with similar functionality into a single category. However, to perform manipulation tasks, it is crucial to identify not only \textit{what} the object's affordances are but also \textit{where} a manipulator should manipulate the object with those affordances, as well as \textit{how} the actions should be performed by the manipulator associated with those affordances. For instance, when using a manipulator to cut a sausage, we need to consider not only the affordances of "cut" and "grasp" but also specific details such as the position and orientation for grasping, as well as the cutting contact point position and direction. In short, affordance semantic segmentation provides what affordances exist, but not where or how they should be executed.

\begin{figure*}[t]
    \centering
    \includegraphics[width=\linewidth]{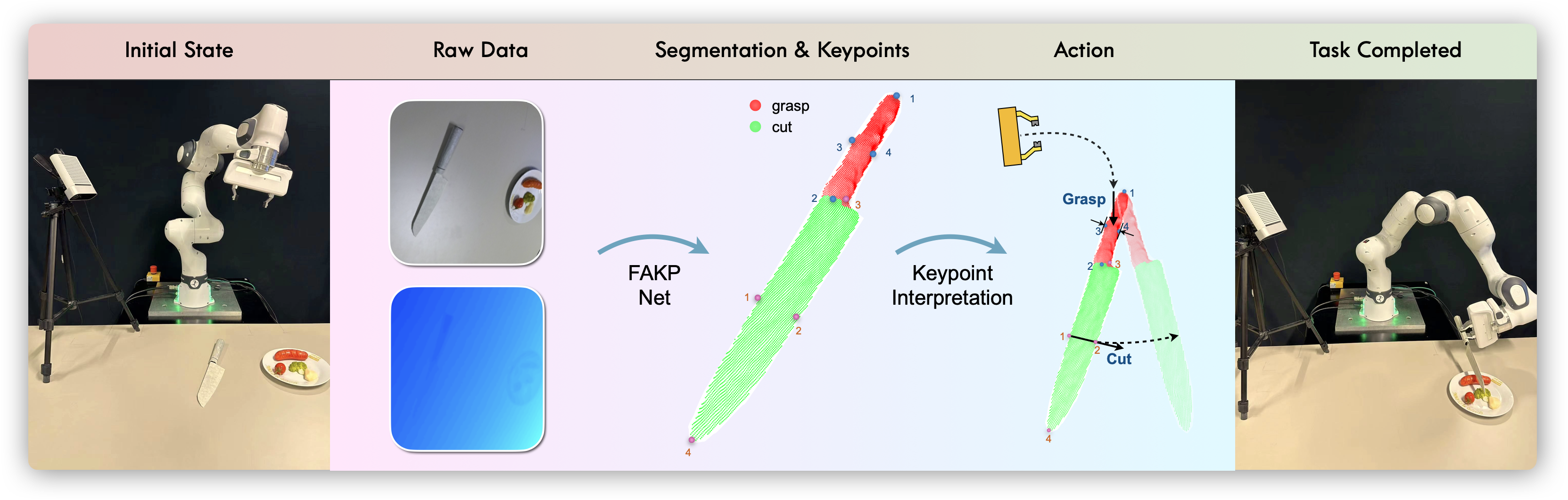} 
    \caption{Affordance-informed Manipulation Pipeline via 3D Keypoints. Each affordance region is represented by four 3D keypoints predicted from RGB-D image by FAKP-Net. Keypoints are interpreted as action position, direction and extent. The manipulator completes the task with provided execution information.}
    \label{teaser}
\end{figure*}

To bridge this gap, \cite{affkp} firstly proposed an action-level approach for object parts, representing objects by pixel-wise affordance labels and 2D image keypoints per affordance. However, the image keypoints fail to capture the object's geometric features and depend heavily on post-processing techniques for application in real-world scenarios. This deficiency in spatial awareness complicates the differentiation of affordances that appear similar, such as the 'contain' and 'scoop' affordances, which, despite their resemblance, differ significantly in size. Moreover, the reliance on 2D keypoints necessitates further post-processing to translate these points into practical applications, especially for tasks involving object manipulation. An example of this limitation is evident in the wrap-grasp action; if the predicted keypoints deviate slightly from the object's surface—even by a few pixels—the action is likely to fail. 

Hence, incorporating both geometric and appearance information into keypoints representation is crucial. It not only helps distinguish between similar affordances but also improves the accuracy of detection for subsequent manipulation tasks. In our work, leveraging this insight, we assign four 3D keypoints to each affordance region, forming the corresponding execution position, direction, and extent, which are explicit representations of where and how to manipulate. For example, as depicted in \autoref{teaser}, when cutting a sausage with a knife, we assign two sets of four 3D keypoints for grasp affordance and cut affordance respectively: for grasping, keypoints 3 and 4 determine the position and orientation of the grasp, and keypoint 2 provides the connection with cut affordance; for cutting, keypoint 3 provides a connection with grasp affordance, and keypoint 2 indicates the contact point of the sharp edge to cut, while the direction from keypoint 1 to keypoint 2 indicates the cutting direction.

In summary, the contribution of this work is twofold. Conceptually, to the best of our knowledge, it is the first to introduce four 3D keypoints to affordance detection, providing a novel representation to guide \textit{where} and \textit{how} a manipulator should engage. Regarding implementation, a novel Fusion-based Affordance Keypoint Detection Network (FAKP-Net) is proposed, which outperforms existing models by significant margins in affordance segmentation task and 3D keypoint detection task on the UMDKP dataset, augmented from UMDGT dataset \cite{affkp}. Real-world experiments of affordance-informed manipulation tasks demonstrate the generalizability of the proposed approach to previously unseen objects.

\section{RELATED WORK}

\noindent\textbf{Affordance Detection:}
Affordance detection has been the subject of numerous studies in the fields of computer vision and robotics \cite{hadjivelichkov_one-shot_2023,chu_learning_2019,sawatzky_weakly_2017, do_affordancenet_2018,gall_adaptive_2017}. \cite{hadjivelichkov_one-shot_2023} finds corresponding affordances both for intra- and inter-class one-shot part segmentation. \cite{do_affordancenet_2018} introduced AffordanceNet, a learning-based end-to-end approach to accurately identify affordances for objects. \cite{chu_learning_2019} focused on learning affordance segmentation for real-world robotic manipulation using synthetic images, contributing insights into enhancing robotic manipulation tasks. \cite{sawatzky_weakly_2017} presented an adaptive binarization technique for weakly supervised affordance segmentation, contributing to the development of efficient detection methods. However, these works only output affordance labels on images, representing what the potential use of objects is.


\noindent\textbf{Keypoint for Manipulation:}
The main applications of keypoints are to cast articulated human parts to joints, thus framing human pose estimation as a joint detection task. In the context of manipulation, we hypothesize 3D keypoints is able to provide the position and geometric details for manipulation tasks. Keypoint representations have been utilized in previous works \cite{manuelli_kpam_2019,gao_kpam-sc_2019,qin_keto_2019,affkp} to tackle the problem at various levels, including the task-level, category-level, and action-level. The KETO framework \cite{qin_keto_2019} is only able to predict a set of task-specific keypoints in a simulation environment, which is not enough to fully specify the robot's action, especially in the real-world scenarios with diverse task contexts and diverse object geometries. Category-level methods\cite{manuelli_kpam_2019,gao_kpam-sc_2019} utilize a set of 3D keypoints to represent objects. However, these category-level methods are object-dependent, and they greatly benefit from prior information about the object's shape, especially when it comes to shape completion \cite{gao_kpam-sc_2019}.


\begin{figure*}[t]
    \centering
    \includegraphics[width=\linewidth]{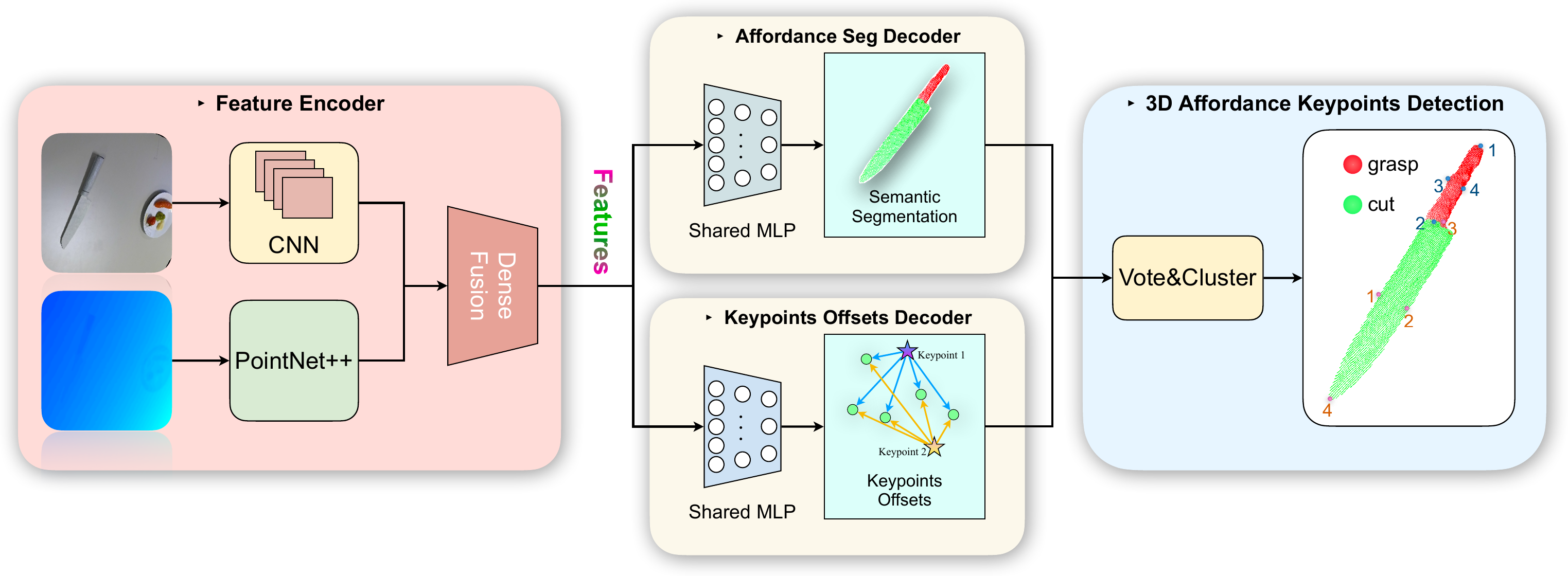} 
    \caption{Overview of FAKP-Net. The feature encoder processes an RGB-D image to extract per-point features. These features are then fed into affordance segmentation decoder and keypoints offsets decoder respectively, predicting per-point semantic labels and per-point translation offsets relative to keypoints. A clustering algorithm is then used to distinguish different affordance regions with the same semantic labels, and points within the same affordance region vote for their keypoints.}
    \label{network}
\end{figure*}

\section{METHOD}
The main challenge we aim to address is the simultaneous detection of point-wise affordances and their corresponding 3D keypoints from RGB-D images. To tackle this task, we propose a network with an encoder-decoder architecture called Fusion-based Affordance Keypoint Network (FAKP-Net), which is described in \autoref{network}. Specifically, given an RGB-D image as input, our network employs a feature extraction module to fuse appearance features and geometry features. The learned features are then passed through an affordance segmentation module and a 3D keypoint detection module respectively. The 3D keypoint detection module is trained to predict per-point offsets relative to keypoints, while the affordance segmentation module is trained to predict per-point semantic labels for object parts corresponding to affordances. Using the learned per-point offsets and labels, we apply the clustering algorithm \cite{comaniciu_mean_2002} to predict a set of four 3D keypoints for each affordance region. These keypoints are utilized to interpret the position and direction information related to the affordance and how it can be effectively executed. For example, as illustrated in \autoref{network}, there are two sets of keypoints associated with the grasp and cut affordances of a knife. For the grasp affordance, keypoints 3 and 4 represent the expected contact points for grasping, while keypoints 2 serve as a connection point with cut affordance. Similarly, for the cut affordance, the direction from keypoint 1 to 2 forms the operating direction, with keypoint 2 also indicating the expected contact point for cut execution. Keypoints 3 in this case also serve as a connection point to connect with grasp affordance.

\noindent\textbf{Datasets and Training:}
The primary source of our data is the UMD dataset \cite{myers_affordance_2015}. This was expanded as UMDGT dataset \cite{affkp}, which added five keypoints for each affordance region in RGB-D images. Within the UMDGT dataset, the support affordance was excluded due to its inherent vagueness. Essentially, any object with a surface could be deemed supportive, blurring its identification as a primary affordance. Given tables are commonly used in manipulation tasks, there's a risk of misidentifying them with a support affordance, confusing them with background categories. Consequently, the UMDGT dataset offers six affordances: grasp, cut, scoop, contain, pound, and wrap-grasp. These images were obtained using a Kinect camera for the original UMD objects \cite{myers_affordance_2015}. Within the five 2D keypoints, the center one often isn't on the object surface, impacting the uptake of geometric features. Moreover, this center point can be inferred from the other keypoints. Therefore, we enhanced the UMDGT dataset by removing the center keypoint and repositioning the others on the object surface. We maintained the initial training-test split from the UMD dataset to ensure consistent comparison in affordance segmentation. 

\noindent\textbf{Feature Encoder:}As shown in \autoref{network}, the feature encoder consists of two components: a PSPNet \cite{zhao_pyramid_2017} with a pretrained ResNet34 \cite{he_deep_2016-1} on the ImageNet dataset \cite{deng_imagenet_2009} for extracting appearance information from RGB images, and a PointNet++ \cite{qi_pointnet_2017} for extracting geometric information from point clouds and normal maps. These two components are then fused together using a DenseFusion block \cite{wang_densefusion_2019} to acquire a fused feature representation for each seed point. After the feature encoder block, each point $p_i$ is associated with a corresponding feature $f_i$.



\noindent\textbf{Affordance Segmentation Decoder:} In our formulation of the 3D affordance keypoints problem, we utilize two learning modules for each visible seed point, one for semantic labels and one for translation offsets to keypoints. We utilize shared Multi-Layer Perceptrons (MLPs) for both decoders and train them jointly using a multi-task loss.

Specifically, with the per-point extracted feature from the encoder, the affordance segmentation decoder predicts the per-point affordance labels. Focal Loss\cite{lin_focal_nodate} is applied for the training:
\begin{equation}
 \mathcal{L}_{semantic }= -\alpha\left(1-c_i \cdot l_i\right)^\gamma \log \left(c_i \cdot l_i\right) 
\end{equation}
where $\alpha$ represents the balance parameter, $\gamma$ denotes the focusing parameter, $c_i$ is the predicted confidence that the $i_{th}$ point belongs to each affordance category, and $l_i$ is the one-hot representation of the ground truth affordance label. In our context, we designate seven labels; label zero is assigned for the background, while the other six are assigned for different affordance categories. We employ $\gamma =2$ and $\alpha = (0.03, 0.12, 0.17, 0.21, 0.17, 0.2, 0.1)$ across all experiments.

\noindent\textbf{3D Keypoints Offsets Decoder:} As depicted in \autoref{network}, given the per-point extracted feature from feature encoder, a 3D keypoints offsets decoder predicts per-point Euclidean translation offset from visible points to four target keypoints. The coordinates of visible seed points, combined with the predicted per-point translation offsets, yield possible keypoint quadruplets—this is the voting process. Within a specific affordance region, the voted points are subsequently collected through clustering algorithms, and these clusters' centers are chosen as the voted keypoints. 

Specifically, given input of visible points ${\{p_i\}}^{N}_{i=1}$ and a keypoint quadruplet ${\{kp_j\}}^{M=4}_{j=1}$ within a specific affordance region $R_{aff, aff \in \{'grasp','cut','scoop','contain','pound','w-grasp'\}}$ , we represent $p_i = [x_i;f_i]$ where $x_i$ refers to the xyz coordinates of seed point and $f_i$ refers to the extracted feature, and $kp_j = [y_i]$ where $y_i$ refers to the 3D coordinates of keypoint. Keypoints offsets decoder processes features $f_i$ to output translation offset $\{of_i^j\}_{j=1}^{M=4}$ for each seed point, where $\{of_i^j\}$ represents translation offsets from $i_{th}$ visible seed point to the $j_{th}$ keypoint. The voted points can then be represented as $vkp_{i}^{j} = x_i + of_{i}^{j}$. To optimize $of_{i}^{j}$, we employ an L1 loss function:
\begin{equation}
\mathcal{L}_{keypoints} = \frac{1}{N} \Sigma_{i=1}^{N}\Sigma_{j=1}^{M}||of_i^j-of_i^{j*}||^{\amalg(p_i\in R_{aff})}
\end{equation}
with $of_i^{j*}$, the ground truth of translation offsets; $of_i^{j}$, the predicted translation offsets; $M=4$, the number of target keypoints for one affordance region; $N$, the total number of seed points; $\amalg$, the indicator function equal to 1 only when point $p_i$ belonging to affordance region $R_{aff}$, and 0 otherwise.

\noindent\textbf{Multi-task Loss:} To jointly supervise the learning of per-point affordance semantic labels and per-point keypoints offsets, a multi-tasks loss is applied by a loss weighting $\lambda$ ($\lambda = 100$):
\begin{equation}
\mathcal{L}_{multi-task} = \mathcal{L}_{keypoints} + \lambda \mathcal{L}_{semantic }
\end{equation}

\begin{figure}[t]
    \centering
    \includegraphics[height=0.325\linewidth]{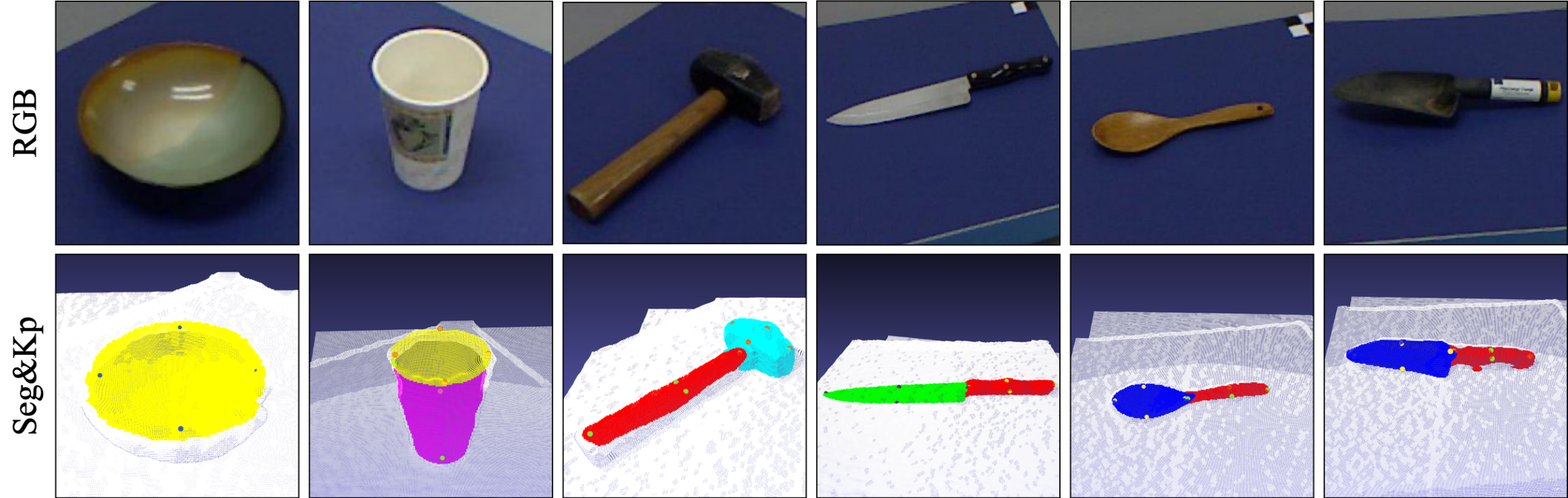}
    \caption{Visualization of Affordance Segmentation and 3D numbered keypoints on UMDKP test dataset. The color coding for the affordance categories is as follows: white for the background; yellow for contain; purple for w-grasp; red for grasp; cyan for pound; green for cut; blue for scoop. Four numbered 3D keypoints attach with each affordance region.}
    \label{testVis}
\end{figure}

\section{EVALUATION}

\noindent\textbf{Affordance Segmentation}: The evaluation of affordance segmentation requires a comparison between valued probability masks and binary labels of ground truth for each class. We employ the $F_\beta^\omega$ metric \cite{margolin_how_2014} to evaluate these masks:
\begin{equation}
    F_\beta^\omega=\left(1+\beta^2\right) \frac{P^\omega \cdot R^\omega}{\beta^2 \cdot P^\omega+R^\omega} .
    \label{fmeasure}
\end{equation}
where $P^\omega$ denotes weighted precision and $R^\omega$ denotes weighted recall, with $\sigma^2 = 5$, $\alpha = \frac{ln0.5}{5}$, and $\beta =1$. Higher weights are assigned to points closer to the ground-truth foreground.

\noindent\textbf{3D Affordance Keypoints}: Normalized Mean Squared Error (NMSE) and Percentage of Correct 3D Keypoints(PCK3D) are commonly used evaluation metrics for keypoints in human pose estimation. Similarly, for each  $aff,aff \in \{'grasp','cut','scoop','contain','pound','w-grasp'\}$, we employ these two metrics to evaluate 3D affordance keypoints.

\begin{equation}
NMSE_{aff} = \frac{1}{N_{aff}M} \Sigma_{i=1}^{N_{aff}}\Sigma_{j=1}^{M} \frac{ ||kp_i^j-kp_i^{j*}||_2}{d_{aff}}
\label{equ:nmse}
\end{equation}

where $kp_i^{j*}$ denotes the ground-truth keypoints; $kp_i^{j}$ denotes the predicted keypoints; $M =4$ denotes the number of target keypoints for each affordance region; $N_{aff}$ denotes the total number of affordance regions for each $aff$ in the test dataset.

The normalized factor $d_{aff}$:

\begin{equation}
\begin{aligned}
d_{aff} &= \frac{1}{N_{aff}M} \sum_{i=1}^{N_{aff}}\sum_{j=1}^{M} \left\|kp_i^{j*}-\bar{kp_{i}}\right\|_2, \\
\bar{kp_{i}} &= \frac{1}{M} \sum_{j=1}^{M} kp_i^{j*}
\end{aligned}
\end{equation}

The normalized factor $d_{aff}$ is also utilized as the threshold of $PCK3D$

\begin{equation}
    PCK_{aff}@0.3 = \frac{C_{0.3d_{aff}}}{C_{aff}}
    \label{equ:pck}
\end{equation}

where ${0.3d_{aff}}$ determines the maximum allowable distance between the predicted keypoint and the ground truth keypoint to be considered correct; $C_{0.3d_{aff}}$ counts the number of correct keypoint quadruplets; $C_{aff}$ counts the total number of ground-truth keypoint quadruplets. The keypoints that are not predicted and misclassified are regarded as incorrect keypoints.

\begin{table}[t]
    \caption{Affordance Segmentation Performance}
    \label{tab:aff}
    \resizebox{\linewidth}{!}{
        \begin{tabular}{|l|c|c|c|c|c|c|c|}
            \Xhline{2\arrayrulewidth}
            \multirow{3}{*}{Affordance} & \multicolumn{7}{c|}{Weighted F-measures} \\
            \cline{2-8} 
            & \multicolumn{2}{c|}{Region-proposal} & \multicolumn{5}{c|}{Encoder-Decoder} \\
            \cline{2-8}
            & AffNet & SRF & AffCorrs & ED-RGB & DeepLab & AffKP* & Ours \\
            \hline
            grasp & 0.73 & 0.31 & 0.65 & 0.72 & 0.62 & 0.72 & \textbf{0.74} \\
            \hline
            cut & 0.76 & 0.41 & 0.81 & 0.74 & 0.60 & \textbf{0.89} & \textbf{0.89} \\
            \hline
            scoop & 0.79 & 0.48 & 0.81 & 0.74 & 0.80 & 0.78 & \textbf{0.83} \\
            \hline
            contain & 0.83 & 0.64 & 0.87 & 0.82 & 0.90 & 0.86 & \textbf{0.92} \\
            \hline
            pound & 0.84 & 0.67 & 0.87 & 0.81 & \textbf{0.88} & 0.79 & \textbf{0.88} \\
            \hline
            w-grasp & 0.81 & 0.26 & 0.89 & 0.77 & 0.73 & 0.82 & \textbf{0.92} \\
            \hline
            \hline
            average & 0.80 & 0.46 & 0.82 & 0.77 & 0.76 & 0.81 & \textbf{0.86} \\
            \Xhline{2\arrayrulewidth}
        \end{tabular}
    }
\end{table}

\begin{table}[t]
    \caption{Affordance-associated 3D Keypoint Performance}
    \label{tab:kp}
    \resizebox{\linewidth}{!}{
        \begin{tabular}{|l|c|c|c|c|c|c|c|}
            \Xhline{2\arrayrulewidth}
            Affordance & grasp & cut & scoop & contain & pound & w-grasp & mean \\
            \hline

            AffKP NMSE & 0.37 & 0.26 & 0.33 & 0.42 & 0.82 & 0.31 & 0.42 \\
            \hline
            AffKP PCK@0.3 & 56.37 & 65.96 & 59.78 & 58.49 & 24.52 & 71.06 & 56.03 \\
            \hline
            \hline
            FAKP NMSE & 0.39 & 0.27 & 0.28 & 0.18 & 0.35 & 0.16 & 0.27 \\
            \hline
            FAKP PCK@0.3 & 42.91 & 64.07 & 66.06 & 82.96 & 52.57 & 97.10 & 67.61 \\

            \Xhline{2\arrayrulewidth}
        \end{tabular}
    }
\end{table}


\section{RESULTS}

\begin{figure}[t]
    \centering
    \resizebox{0.9\linewidth}{!}{%
        \includegraphics{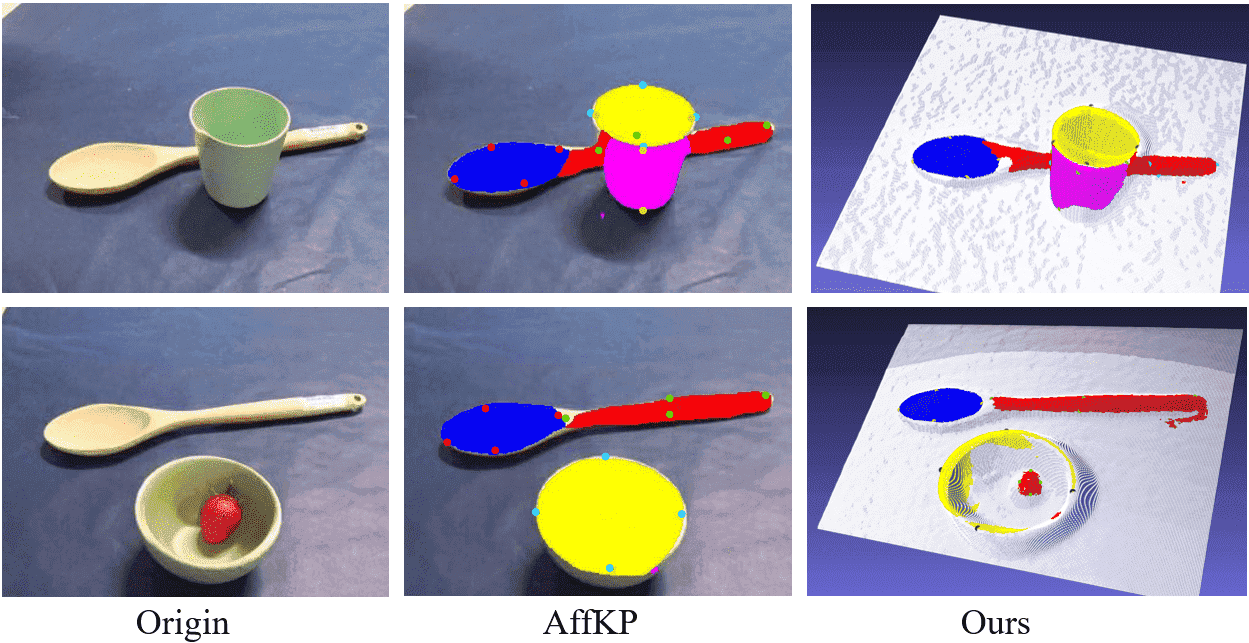}
    }
    \caption{Outputs of FAKP and AffKp on previously unseen objects in real-world (objects from IKEA).}
    \label{fig:clutter}
\end{figure}

In this section, we discuss the performance of FAKP-Net based on the metrics of affordance segmentation and associated 3D keypoint detection. Visualizations of FAKP-Net outputs on the UMDKP test dataset in \autoref{testVis} show that FAKP-Net can simultaneously predict affordance semantic labels and per-affordance 3D keypoints. The performance of affordance segmentation is summarized in \autoref{tab:aff} and the performance of 3D keypoints is shown in \autoref{tab:kp} and \autoref{fig:clutter}. 

\noindent\textbf{Affordance Segmentation Performance:}
 In \autoref{tab:aff}, this evaluation includes a comparison with several baseline methods: AffordanceNet(AffNet)\cite{do_affordancenet_2018}, SRF\cite{myers_affordance_2015}, AffCorrs\cite{hadjivelichkov_one-shot_2023}, ED-RGB\cite{nguyen_detecting_2016}, DeepLab\cite{chen_deeplab_2017} and AffKP\cite{affkp}. Except for AffKP, others follow a same train-test split rule as UMD dataset. For a fair comparison, we retrained AffKP and tested it with same split rule. The first two methods utilize region-proposal architectures, whereas the last four are based on simpler encoder-decoder structures. Those region-proposal methods simplify the problem by pre-processing the image to obtain regions of interest (RoIs) to get better performance. Nonetheless, as indicated by the results in \autoref{tab:aff}, our proposed FAKP-Net significantly outperforms not only existing encoder-decoder methods but also region-proposal methods, achieving state-of-the-art results across all affordance categories. Of particular note is the superior performance of our method in the wrap-grasp affordance category, where we surpass other methods by a margin of at least 10\%. This can be attributed to the heightened sensitivity of the w-grasp affordance to geometric features, as opposed to appearance features. Our FAKP-Net is capable of effectively capturing these geometric features, which sets it apart from other methods. For instance, a mug can have complex patterns printed on its surface, leading to complicated appearance features and potentially hindering the detection of wrap-grasp affordance regions. However, when geometric information such as the size and shape of the mug is fused with the appearance features, the detection of wrap-grasp affordance becomes more straightforward.

\noindent\textbf{Affordance-associated 3D Keypoints Performance:}
In addition to affordance segmentation, the FAKP-Net concurrently generates 3D keypoints for each identified affordance region. For the evaluation of our predicted 3D keypoints, we employ metrics NMSE (Normalized Mean Squared Error) and PCK3D (Percentage of Correct 3D Keypoints). Notably, AffKp outputs 2D keypoints for each affordance region, and to fairly compare with our 3D affordance keypoints, we use the same way as how we get the point clouds to cast the 2D keypoints generated from AffKp to 3D keypoints. The results is shown in \autoref{tab:kp}. The NMSE score quantifies the average error, normalized with respect to the constant $d_{aff}$, which is specific to each affordance class. This constant reflects the average distance of the four keypoints from their mean center point. The PCK3D score quantifies the percentage of correct keypoints. 

As in \autoref{tab:kp}, FAKP exhibits varying performance across six distinct affordances. The wrap-grasp and contain affordances both demonstrate low NMSE scores and high PCK@0.3. Notably, they both also exhibit high weighted F-measures at 0.92, indicating successful capture of their respective appearance and geometric features. The scoop and cut affordances demonstrate relatively moderate performance. In contrast, the grasp and pound affordances exhibit poorer performance in terms of weighted F-measures, NMSE, and PCK@0.3. This poorer performance can be attributed to their inherently ambiguous nature. For instance, both the handle and the head of a hammer are black and graspable, and distinguishing features between the handle and the head of the hammer are not always clear.

Compared with AffKP, FAKP showcases enhanced performance, with a notable increase of 11.58\% in PCK@0.3 and a reduction of 0.15 in NMSE. This improvement is especially pronounced in the affordances related to contain, pound, and w-grasp, which are more susceptible to geometric discrepancies compared to grasp, cut, and scoop. A case in point is the affordance for contain, which can be easily mistaken for the scoop due to their visually similar circular shapes, albeit with different sizes. This underscores the importance of integrating geometric feature fusion in our approach to detecting affordance keypoints. The comparative visualizations of FAKP and AffKP, as shown in \autoref{fig:clutter}, highlight capability of FAKP to navigate complex scenarios effectively. For example, in scenario of first row, the spoon's handle, obscured by a mug, leads to erroneous detection of a grasp keypoint within the contain area by AffKp, mistaking part of the contain affordance for grasp. In another scenario of row two, featuring a tomato in a bowl, AffKP fails to detect the tomato as graspable. This limitation stems from its reliance solely on appearance features, which are insufficient for distinguishing the tomato from a pattern on the bowl. The \autoref{tab:kp} and \autoref{fig:clutter} demonstrate the superiority of FAKP compared with AffKp from accuracy and visualization.

\section{EXPERIMENTS}
In this work, we follow the experiment pipeline in \cite{affkp} to showcase the reliability of our method in accomplishing manipulation tasks with previously unseen objects. We manually interpret the network outputs into real-world robot's actions. The outputs as a representation of \textit{what},\textit{where} and \textit{how} details of affordances contribute to manipulation planning and execution.

\noindent\textbf{Set-up in the Real-world:}
For our real-world experiments, as shown in \autoref{teaser}, we use a Franka Emika 7-DoF robotic arm fitted with a 2-fingered gripper and the robot's maximum payload of 3kg. A Kinect Azure camera captures the RGB-D image from the workspace and camera parameters are calibrated before experiments. To test the generalizability of our network, all objects in real-world experiments are unseen before and sourced from IKEA.

\begin{figure}[t]
    \centering
    \resizebox{0.9\linewidth}{!}{%
        \includegraphics{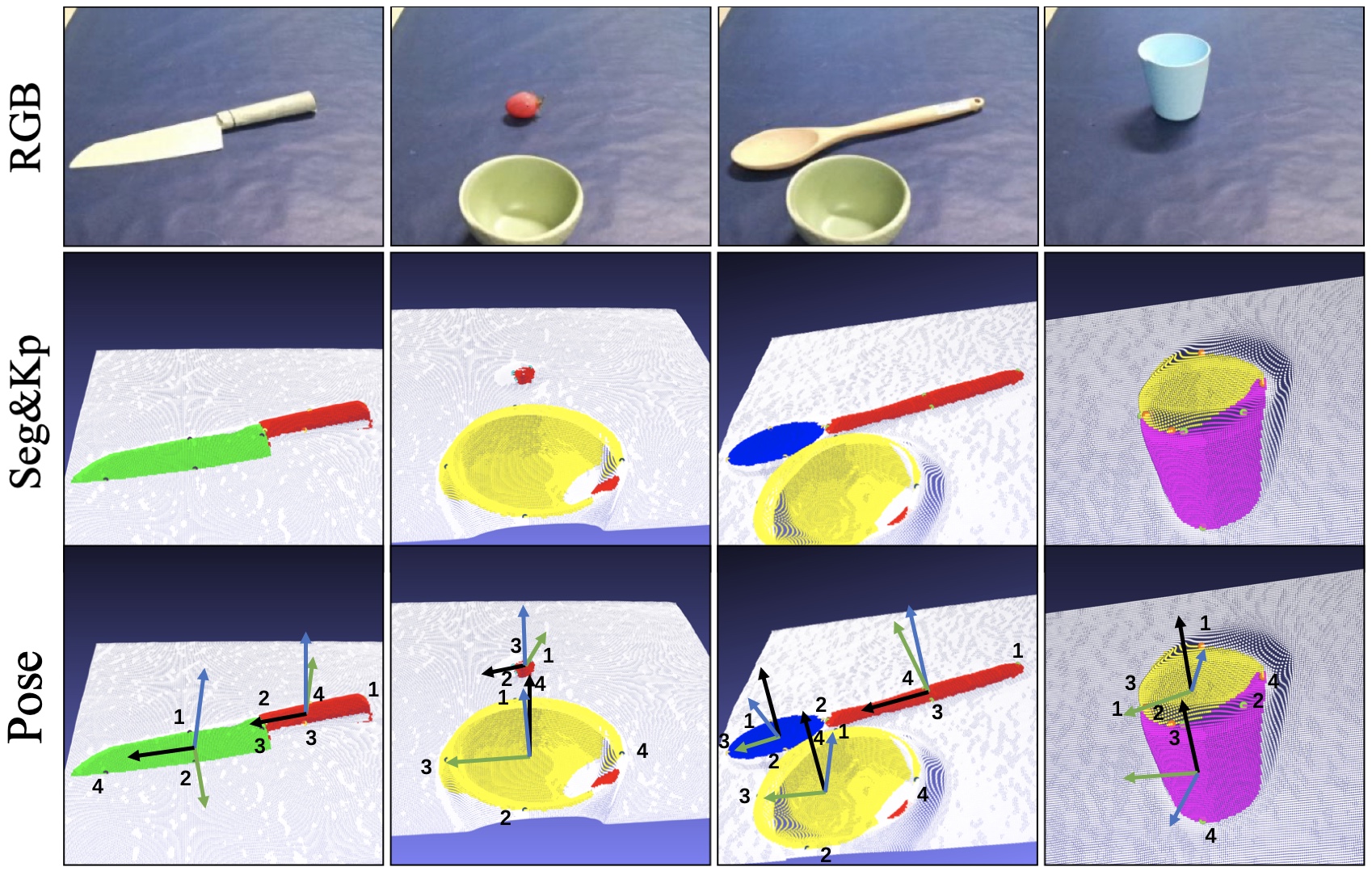}
    }
    \caption{Visualization of Affordance Segmentation and 3D keypoints quadruplet on previously unseen objects in real-world (objects from IKEA).}
    \label{fig:exp_vis}
\end{figure}

\noindent\textbf{Keypoint Interpretation:}
We mainly employ keypoints quadruplet and the center point of it to determine two axes (x, y) and the origin of the frame. We use black, green and blue axes to represent the x, y and z axes, as shown in \autoref{fig:exp_vis}. In order to avoid confusion, the x-axis always represents the principal axis of the object part. However, the axis required for different tasks won't always be the x-axis.
\begin{itemize}
    \item \textbf{Grasp:} The origin of the frame is determined by the center point, which is the average coordinate of the 4 keypoints. The task-dependent axis which is the y-axis is computed by keypoints 3 and 4. Then the x-axis is computed by y-axis and keypoints 1 and 2. Lastly, the z-axis is computed by cross-product.
    \item \textbf{Contain and scoop:} The origin of the frame is computed by the average coordinate of the 4 keypoints. The y-axis is computed by keypoints 3 and 4. Then the z-axis is computed by y-axis and keypoints 1 and 2. Lastly, the x-axis is computed by cross-product.
    \item \textbf{Wrap-grasp:} The origin of the frame is determined by the average coordinate of keypoint 3 and 4. The task-dependent axis which is the y-axis is computed by keypoints 1 and 2. Then the x-axis is computed by y-axis and keypoints 3 and 4. Lastly, the z-axis is computed by cross-product.
    \item \textbf{Cut:} For cut affordance, the y-axis, which indicates the operational direction, is computed by the keypoints 1 and 2. Similarly, the x-axis is computed by both y-axis and keypoints 3 and 4. Lastly, the z-axis is computed by cross-product.
\end{itemize}

\noindent\textbf{Manipulation Task Specifications:}
Four affordance-informed manipulation tasks are introduced to test our pipeline. The video demo is in supplementary materials.

\begin{table}[t]
    \caption{Robot Experiment Statistics}
    \label{tab:exp_statistic}
    \centering
    \begin{tabular}{|>{\centering\arraybackslash}m{4em}|>{\centering\arraybackslash}m{2.5em}|>{\centering\arraybackslash}m{3em}|>{\centering\arraybackslash}m{4em}|>{\centering\arraybackslash}m{3em}|>{\centering\arraybackslash}m{4em}|}
        \Xhline{2\arrayrulewidth}
        \textbf{Task} & \textbf{\scriptsize \#Trails} & \textbf{\scriptsize \#Failure} & \textbf{\scriptsize \#Planning Failure} & \textbf{\scriptsize \#Grasp Failure} & \textbf{\scriptsize \#Execution Failure} \\
        \hline 
        1 & 30 & 1 & 1 & 0 & 0 \\
        \hline 
        2 & 30 & 4 & 2 & 2 & 0 \\
        \hline 
        3 & 30 & 10 & 2 & 3 & 5 \\
        \hline 
        4 & 30 & 6 & 4 & 0 & 2 \\
        \hline
    \end{tabular}
\end{table}

\begin{enumerate}
    \item \textbf{Put a tomato into a bowl:} The first manipulation task aims to evaluate the contain affordance. The objective is to place a tomato into a bowl. After grasping the tomato, the desired position is determined by the origin of the contain affordance. The gripper then releases the tomato above this target position. The task is considered successful if the tomato is contained within the bowl.
    \item \textbf{Cut sausage with a Knife:} The second manipulation task focuses on assessing the grasping and cutting affordance. The goal is to first grasp the knife and then use its blade to make contact with the sausage. The grasp position and orientation are determined by the origin and y-axis of grasp affordance. Keypoint 2 of the cut affordance serves as the contact point between the knife and the sausage, while the direction of the y-axis indicates the cutting direction. If the sharp side of the blade makes contact with the sausage, the task is considered successful.
    \item \textbf{Scoop from a bowl:} The third manipulation task tests the grasp, scoop, and container affordance. Initially, the gripper grasps the tool. Subsequently, the y-axis of the scoop affordance is aligned with the y-axis of the contain affordance. Success is achieved if the tool can enter and exit the bowl without colliding with its edges.
    \item \textbf{Wrap-grasp a cup:} The final manipulation task examines the wrap-grasp affordance. The objective is to wrap-grasp, lift, and hold a cup. The desired position for this task is determined by the origin of the wrap-grasp affordance, while the grasp width is determined by the distance between keypoint 1 and 2 and grasp orientation is determined by the y-axis of wrap-grasp affordance. The task is considered successful if the manipulator can hold the cup in the air for a duration of 3 seconds.
\end{enumerate}

\noindent\textbf{Manipulation Tasks Results:}
The visualization results from captured RGB-D image of the workspace are shown in \autoref{fig:exp_vis}, while the statistics for the four distinct manipulation tasks are summarized in \autoref{tab:exp_statistic}. Most fail cases are from failed motion planning of motion planners which can be trapped in local minima due to its reliance on non-convex optimization. This issue could potentially be mitigated by employing sampling-based motion planners. Grasp failure in \autoref{tab:exp_statistic} means the robot was unable to successfully grasp the object, or excessive relative motion between the gripper and the grasped object occurred. This issue could potentially be addressed by re-perceiving on-hand objects. The execution failure pertains to scenarios where the robot collides with an object despite successful planning, or the success requirements for the respective tasks are not met.

\section{LIMITATIONS}
We believe that the limitations of our approach are mainly from the dataset and the hardware:

\noindent\textbf{Dataset:} Since our UMDKP dataset is enhanced from UMD dataset. UMDKP dataset has limited diversity in terms of object categories, scenes, and environments. This can restrict the generalizability of models trained on the dataset to real-world scenarios. For example, UMD dataset only includes limited affordance categories; the background is monotonous; there are no cluttered scenes.

\noindent\textbf{Hardware:} This limitation is mainly from Kinect camera and Franka Emika manipulator. Kinect Azure camera used in our real-world experiments may have difficulties capturing reflective objects and black objects, because it uses infrared depth sensors to capture depth information, and infrared sensors may struggle with reflective surfaces and absorbent black objects. The gripper and manipulator specifications also restrict the generalizability in real-world experiments. For example, the manipulator has difficulty manipulating the objects which exceed the 3Kg maximum payload of the manipulator; the two-finger gripper also restricts the manipulation tasks, for example, sliding when grasping the hammer owing to its uneven weight distribution nature as our video demo shows. 

\section{CONCLUSION}
To address downstream planning and action challenges associated with affordances, we introduced the Fusion-based Affordance Keypoint Network(FAKP-Net), which can output affordance-associated 3D keypoints to support affordance-informed manipulation tasks. Real-world manipulation experiments showcase the reliability of our method in accomplishing manipulation tasks with never-before-seen objects. Future work will explore FAKP-Net in a more complex environment. All codes and data will be public.

\section*{ACKNOWLEDGMENT}

This research is supported by the Agency for Science, Technology and Research (A*STAR) under its AME Programmatic Funding Scheme (Project \#A18A2b0046).




\clearpage


\bibliographystyle{IEEEtran}
\bibliography{example}

\end{document}